# A Reproducible, License-Aware Distillation Recipe for CPU-Deployable Safety Classification

**Edson Rodrigues da Cruz Filho[1,2], Paulo Ricardo Ferreira Neves[1,3], Paulo Henrique Eleuterio Falsetti[1,4], João Vitor Pavan[1], Ian Degaspari[1], Henrique Vieira Laturrague[1], Patrick Vieira Laturrague[1], Guilherme Nielsen Dias[1], Marccello Wilson Perez Berto[1], Gustavo Voltani Von Atzingen[1,2]**

[1] Quickium Technology Ltd., Piracicaba, São Paulo, Brazil.
[2] Federal Institute of Education, Science and Technology of São Paulo (IFSP), Piracicaba Campus, São Paulo, Brazil.
[3] University of São Paulo (USP), Escola Superior de Agricultura Luiz de Queiroz, Piracicaba Campus, São Paulo, Brazil.
[4] Federal University of São Carlos (UFSCar), Sorocaba Campus, São Paulo, Brazil.

Corresponding author: Edson Rodrigues da Cruz Filho, efilho@quickium.com, ORCID 0009-0007-4112-6100.

## ABSTRACT

Deploying a safety layer for large language models on commodity hardware is constrained by the guards available to do it: current open guard models hold between 1 and 9 billion parameters, are oriented toward the graphics processing unit, and answer in seconds per request on a central processing unit. This paper presents a reproducible, license-aware knowledge-distillation recipe addressing that constraint. A strong open guard labels a corpus of roughly 97,000 prompts, drawn from 24 public datasets, into seven safety categories aligned to a public hazard taxonomy, and a fleet of small students spanning lexical, shallow, encoder and generative architectures is trained to reproduce that signal. The corpus is partitioned at the license boundary, so that a deployable and a research model differ only in their training data and the cost of that restriction becomes measurable. Every model is scored against an independent gold benchmark of 6,361 rows over four slices, labeled apart from the teacher and including a slice of harmless prompts that makes over-defense measurable. The distilled students match the teachers on adversarial text within overlapping confidence intervals and reduce false alarms on harmless prompts, the smallest generative student reaching 3.8% against 4.8% for the 8-billion-parameter teacher, while the encoder classifies in roughly 24 ms per request on CPU. Per-class rebalancing is the only decisive ingredient of the recipe. No superiority over the distilled guards is claimed; on the clean reference slice they remain ahead.

## 1 Introduction

Large language models (LLMs) deployed in user-facing systems require a safety layer that decides whether a prompt or a response violates a safety policy (Dong et al., 2025). The dominant approach is a dedicated guard model, an instruction-tuned LLM that reads a hazard taxonomy in context and emits a verdict together with the categories it judges to be violated (Inan et al., 2023). Such guards set the quality standard for open safety classification, but they carry a deployment cost: current models span between 1 and 9 billion parameters (Zeng et al., 2024; Fedorov et al., 2024), are oriented toward the graphics processing unit (GPU), and answer in seconds per request when forced onto a central processing unit (CPU) (Chrapek et al., 2025).

That deployment cost is the problem this work addresses. Many settings require the safety layer to run locally on commodity CPU hardware, whether for reasons of cost, latency, or data locality, and there a multi-billion-parameter GPU guard is impractical, because the hardware is unavailable and a per-request latency of several seconds does not scale (Fedorov et al., 2024). What such settings need is a classifier of comparable quality that runs on CPU within a fraction of that time.

The literature offers two families of alternatives, neither of which resolves this situation on its own. Large open guard models provide broad coverage but remain GPU-bound. Lightweight encoders such as toxicity classifiers run cheaply on CPU but cover only a narrow slice of a multi-hazard policy (Zaratiana et al., 2026). Safety benchmarks, in turn, are typically consumed to score a model rather than paired with an independent gold set, one that is labeled apart from the system under test and that includes a benign slice (Zhang et al., 2026). Without such a benign slice, over-defense, the blocking of harmless prompts, cannot be measured (Varshney et al., 2024).

This paper addresses that gap with a safety classifier that runs on CPU and is produced under an auditable recipe. A strong open guard, Llama Guard 3 8B (Inan et al., 2023), labels a corpus assembled from public sources into a seven-category taxonomy aligned with MLCommons AILuminate v1.0 (Ghosh et al., 2025), and a fleet of small students distills that signal. The study evaluates the resulting models against an independent gold benchmark and reports the distillation gap under identical conditions. The evidence indicates that distillation transfers the teacher's signal into students one to two orders of magnitude smaller and runnable on CPU (Fedorov et al., 2024; Guo et al., 2025), which match the teacher on adversarial text and reduce over-defense, while the teacher retains an advantage on clean prompts. Consistent with this scope, the work is presented as a resource and benchmark contribution rather than a claim of state-of-the-art performance, and it does not claim to surpass Llama Guard 3.

The contributions are five. The first is a data pipeline that normalizes 24 public datasets into a single schema, partitions them at the license boundary into a commercial and a noncommercial bucket, and enforces a train-to-evaluation leakage gate (Longpre et al., 2024; Sasse et al., 2023). The second is a distillation procedure whose teacher is chosen by a comparative shootout and whose label overrides are auditable (Lan et al., 2025), keeping an evaluation gold set that the teacher never sees. The third is a student fleet spanning lexical, shallow, encoder, and generative architectures, compared for classification quality under a single protocol with bootstrap confidence intervals. The fourth is a set of ablations that isolate which training choice is decisive (J. Li & Kim, 2024). The fifth is a measurement of CPU inference latency across the fleet, which establishes the quality-versus-latency trade-off on defined hardware profiles.

The remainder of the paper is organized as follows. Section 2 reviews prior work on guard models, safety datasets, and lightweight detectors. Section 3 defines the taxonomy and its mapping to the AILuminate hazards. Section 4 describes the data pipeline, and Section 5 the distillation procedure and the student fleet. Section 6 sets out the evaluation protocol; Sections 7 and 8 report the classification results and the CPU latency; and Section 9 discusses limitations before the conclusion in Section 10.

## 2 Related Work

### 2.1 GUARD MODELS

The prevailing design for safety classification is a dedicated guard model that reads a prompt, and optionally a response, and returns a verdict over a hazard taxonomy. Llama Guard (Inan et al., 2023) established the pattern of an instruction-tuned LLM that consumes a swappable taxonomy in context and emits a safe or unsafe judgment with the violated categories; its Llama Guard 3 line, at 8 and 1 billion parameters, represents the current generation and serves in this work as both the distillation teacher and a baseline. Other open guards apply the same design with different base models and interfaces: ShieldGemma (Zeng et al., 2024) is built on Gemma, Granite Guardian (Padhi et al., 2024) ships under the permissive Apache 2.0 license, and WildGuard (S. Han et al., 2024) additionally releases the WildGuardMix corpus that this work reuses as training data. These guards adopt divergent hazard taxonomies, however, so they cannot be scored head to head against the students on the AILuminate-aligned scheme used here, a constraint discussed in Section 9. The trait that is decisive for deployment is shared across the family: current guards fall between 1 and 9 billion parameters (Zeng et al., 2024; Fedorov et al., 2024) and are GPU-bound for usable latency, running at second scale on CPU (Chrapek et al., 2025), which is precisely the setting this work targets.

### 2.2 SAFETY DATASETS, BENCHMARKS, AND STANDARDS

Training and evaluation rest on public safety resources. MLCommons AILuminate v1.0 (Ghosh et al., 2025) supplies a 12-hazard taxonomy scored by a classifier ensemble into letter grades; adopting it as the base taxonomy buys comparability with guards evaluated on the same scale and a citable reference point. Evaluation draws on adversarial and diagnostic suites, among them the adversarial stress set ALERT (Tedeschi et al., 2024), AttaQ (Kour et al., 2023), SALAD-Bench (L. Li et al., 2024), and SimpleSafetyTests (Vidgen et al., 2023). Training draws on red-teaming and preference corpora, including AdvBench (Zou et al., 2023) and HarmBench (Mazeika et al., 2024), which are harmful by construction and motivate never-safe label overrides, together with BeaverTails (Ji et al., 2023), whose noncommercial Creative Commons license illustrates the boundary that the corpus split manages. Further dialogue and instruction sources widen coverage of harm and refusal: the Anthropic harmlessness and red-teaming data (Bai et al., 2022; Ganguli et al., 2022), CoCoNot (Brahman et al., 2024), DiaSafety (Sun et al., 2022), Do-Not-Answer (Wang et al., 2023), ProsocialDialog (Kim et al., 2022), UltraSafety (Guo et al., 2024), the multilingual Aya red-teaming set (Aakanksha et al., 2024), in-the-wild jailbreaks (Jiang et al., 2024), domain-specific medical safety (T. Han et al., 2024), the toxicity corpora ToxicChat (Lin et al., 2023) and Jigsaw (Jigsaw, 2018), and synthetic personally identifiable information (Gretel.ai, 2024). Brazilian-Portuguese resources, ToLD-BR (Leite et al., 2020) and HateBR (Vargas et al., 2022), are reserved under noncommercial terms for a later multilingual milestone. A distinct line establishes over-defense as a first-class failure mode: XSTest (Röttger et al., 2024) and OR-Bench (Cui et al., 2024) show that exaggerated safety is measurable, and systematic evaluation of defense strategies confirms the trade-off against over-defensiveness (Varshney et al., 2024), which legitimizes reporting a benign false-positive rate alongside detection quality.

### 2.3 LIGHTWEIGHT AND DISTILLED DETECTORS

The student side of this work builds on established compression and lightweight-architecture research, a line that has since been applied to guardrails themselves (Fedorov et al., 2024; Guo et al., 2025). Knowledge distillation (Hinton et al., 2015) is the transfer mechanism by which a small student is trained to reproduce a strong teacher's output. On the encoder side, the BERT lineage (Devlin et al., 2019), compressed as DistilBERT (Sanh et al., 2019) and MiniLM (Wang et al., 2020), provides fast single-pass classification, while the convolutional text classifier of Kim (2014) remains a cheap shallow baseline. On the generative side, low-rank adaptation (LoRA) (Hu et al., 2021) permits parameter-efficient fine-tuning of a small instruction model through an adapter that can later be merged into the base weights. Prior single-purpose detectors mark the narrow end of the design space: Detoxify (Hanu, 2020) is a CPU-cheap encoder that scores toxicity but covers only part of a multi-hazard policy, and Presidio (Microsoft, 2018) matches personally identifiable information by entity type. These support the central move of the present work, which is to distill a large guard into a fleet of small students spanning lexical, shallow, encoder, and generative architectures, all cheap enough to run on CPU, rather than to treat any single narrow detector as a full guard.

### 2.4 WHERE THIS WORK DIFFERS

Against this background, the contribution is not a larger guard but a different combination of properties. The recipe is reproducible and license-aware, in a field where dataset licensing is often left unaudited (Longpre et al., 2024): the corpus is drawn from public sources and partitioned at the license boundary, so a shippable model and a research model differ only in the data they consume. The evaluation uses an independent gold benchmark aligned to a public standard, with a benign slice that makes over-defense measurable rather than assumed. The resulting artifact is CPU-deployable, with latency measured on defined hardware profiles instead of estimated. Finally, the study separates the deploy-pick, the best quality-latency student, from the deployed reality, the model actually placed in production, and reports both. Taken together, these occupy the middle ground that the literature leaves open between large GPU-bound guards and narrow single-purpose detectors, and the work is framed as a resource and benchmark contribution rather than a state-of-the-art claim.

## 3 Taxonomy

The classifier emits seven safety categories, denoted C1 through C7. Rather than define categories from scratch, the taxonomy is derived from MLCommons AILuminate v1.0 (Ghosh et al., 2025), whose 12 hazards (S1 through S12) provide a public, citable reference. Grounding the categories in a standard is a deliberate choice: an ad-hoc taxonomy yields verdicts that are neither comparable across systems nor citable, whereas a standardized harm taxonomy makes both possible, an argument that motivates the adoption of a public standard over a bespoke scheme (Weidinger et al., 2021). The seven categories group the 12 hazards so that each class retains enough training support at the scale of the corpus, and for benchmarking against AILuminate each category expands back to its constituent hazard codes, a lossy but consistent transformation. Table 1 lists the mapping.

**Table 1.** Mapping of the seven safety categories to the MLCommons AILuminate v1.0 hazard codes.

| Code | Slug | Scope | AILuminate hazards |
|---|---|---|---|
| C1 | `physical_harm` | Operational instructions for violent acts, and for chemical, biological, radiological, and nuclear (CBRN) or conventional weapons | S1, S9 |
| C2 | `hate_defamation` | Hate speech against identifiable groups; defamation of a real, named person | S5, S10 |
| C3 | `sexual_content` | Sexually explicit content, sexual crimes, and child sexual abuse material (CSAM), the last always flagged at maximum severity | S3, S4, S12 |
| C4 | `self_harm` | Encouragement of, or instructions for, suicide and self-injury | S11 |
| C5 | `sensitive_info` | Disclosure of personally identifiable information (PII), doxxing, and leaks of confidential corporate information | S7, S8, and an extension |
| C6 | `harmful_advice` | Harmful specialized advice (medical, legal, financial) and harm vectored through advice | superset of S6 |

| Code | Slug | Scope | AILuminate hazards |
|---|---|---|---|
| C7 | illicit_acts | Nonviolent crime such as fraud, cybercrime, drug offenses, and money laundering | S2 |

*Source: the authors, aligned to MLCommons AILuminate v1.0 (Ghosh et al., 2025).*

The grouping follows two principles. Categories merge when public training data for an individual hazard is too thin to sustain a class on its own, as with the child-sexual-abuse (S4) and weapons (S9) hazards, or when hazards share detection features, so that violence and weapons form a single physical-harm capability (C1), defamation and hate form an attack on an identifiable target (C2), and privacy and intellectual property form information that should not leak (C5). Three hazards remain singletons because their signals do not combine cleanly with any other, namely nonviolent crime (S2, mapped to C7), specialized advice (S6, mapped to C6), and self-harm (S11, mapped to C4). Self-harm is itself sparse in public data, so keeping it isolated leaves C4 with limited training support, a constraint the discussion revisits.

Two categories extend the standard. C6 is defined as a superset of the AILuminate Specialized Advice hazard (S6): it retains S6 and adds advice-vectored harms that fall outside that frame, the distinction from C7 being advice-shaped guidance as opposed to instructions to commit a crime. C5 extends the privacy hazards (S7 and S8) with confidential corporate disclosure, which no public standard covers; because no public corpus of corporate leaks exists to train that facet separately, it is folded into C5.

The output is multi-label. Each category carries an independent sigmoid activation and a per-category threshold, 0.5 by default, so that the safe verdict is the all-zeros vector and thresholds can be retuned without retraining. A verdict is emitted as a single line: the word safe when no category fires, or the token unsafe followed by the triggered codes and their scores, comma-separated in descending order, with sub-threshold categories omitted. Reporting codes rather than slugs keeps a safe verdict close to one token and a flagged verdict to a few, while the retained scores let downstream policy tune each category independently.

Three areas are deliberately outside the scope of v0. Prompt injection and jailbreak detection is treated as a separate research line, with its own taxonomy and dedicated models. Elections and disinformation are excluded while the corresponding part of the standard, an additional hazard beyond the 12 of v1.0 (S13), remains unstable. Code-interpreter abuse is excluded because the gateway exposes no code execution and therefore no attack surface for it.

## 4 Data Pipeline

The corpus was assembled entirely from public datasets, normalized into a single schema and partitioned according to license. The normalization stage converted 25 public datasets into a common six-column layout holding a row identifier, the prompt text, the originating source, a pre-teacher class hypothesis, the language, and the license bucket. The class hypothesis was derived deterministically from each source's own labels and was never used as a training target; it was retained only as a sanity check against the labels later assigned by the teacher. Of the 25 normalized datasets, 24 entered the corpus: SafetyBench was normalized but excluded, because its multiple-choice format with a withheld answer key does not match the classification task studied here.

Partitioning followed the license boundary at the level of the individual row, so that every prompt belongs to exactly one bucket. The commercial bucket collects sources whose licenses permit commercial use and is the only bucket a deployable model may consume. The noncommercial bucket collects sources under noncommercial or copyleft terms and is therefore restricted to research. A third bucket holds the evaluation benchmarks, which never enter training. Table 2 reports the resulting composition.

**Table 2.** Corpus composition after normalization and license partitioning.

| Bucket | License terms | Sources | Rows | Role |
|---|---|---|---|---|
| `commercial` | permissive, commercial use allowed | 15 | 75,112 | training, deployable recipe |
| `nc` | noncommercial or copyleft | 4 | 22,332 | training, research only |
| `eval` | benchmark, never trained on | 5 | 6,802 | evaluation (gold) |

*Source: the authors, from the corpus release manifest.*

The commercial bucket draws on WildGuardMix (S. Han et al., 2024), the Jigsaw toxic-comment corpus (Jigsaw, 2018), the harmless-base split of the Anthropic human-preference data (Bai et al., 2022) and the associated red-teaming set (Ganguli et al., 2022), the adversarial behavior suites HarmBench (Mazeika et al., 2024) and AdvBench (Zou et al., 2023), the refusal-oriented sets Do-Not-Answer (Wang et al., 2023) and CoCoNot (Brahman et al., 2024), the vanilla portion of WildJailbreak (Jiang et al., 2024), the medical-advice set MedSafetyBench (T. Han et al., 2024), the dialogue-safety corpora DiaSafety (Sun et al., 2022) and ProsocialDialog (Kim et al., 2022), the harmful-instruction set UltraSafety (Guo et al., 2024), the multilingual red-teaming set of Aakanksha et al. (2024), and a synthetic corpus of personally identifiable information (Gretel.ai, 2024). The noncommercial bucket holds BeaverTails (Ji et al., 2023), ToxicChat (Lin et al., 2023), and the Brazilian-Portuguese resources HateBR (Vargas et al., 2022) and ToLD-BR (Leite et al., 2020). The evaluation bucket holds the public AILuminate v1.0 samples (Ghosh et al., 2025), ALERT (Tedeschi et al., 2024), AttaQ (Kour et al., 2023), SALAD-Bench (L. Li et al., 2024), and SimpleSafetyTests (Vidgen et al., 2023).

Two recipes consume the training buckets. The deployable recipe trains on the commercial bucket alone, and the research recipe trains on the union of the commercial and noncommercial buckets. Because both recipes share the same teacher, the same hyperparameters, and the same evaluation hold-out, any difference in their scores is attributable to the training data alone, which turns the license boundary into a measurable quantity rather than an unexamined constraint.

Leakage control operated in two tiers. The mandatory first tier computed an exact hash over a normalized form of each prompt and resolved collisions by a fixed survival priority in which the evaluation bucket outranks the commercial bucket, which in turn outranks the noncommercial bucket. A collision between training and evaluation therefore removes the training copy, since a benchmark prompt leaking into training would inflate the reported scores. A collision between the two training buckets keeps the commercial copy, so that the deployable recipe is never penalized. The optional second tier applied MinHash with locality-sensitive hashing to catch near-duplicates. Any bucket losing more than 1% of its rows to deduplication was flagged for inspection rather than silently reduced.

The benign slice required separate treatment. Because every gold benchmark listed above is unsafe by construction and contains no harmless prompts, none of them can measure false alarms. A fixed set of 600 English benign prompts was therefore reserved into the evaluation bucket during per-source preprocessing under a fixed random seed, drawn from commercial sources that contain harmless text and weighted toward adversarially benign prompts, meaning text that superficially resembles a violation while being harmless. Reserving these rows before the training stage means they never enter training, so there is no leakage to undo afterwards. They retain their original source attribution and a safe label. The full corpus is regenerable from the public sources through the pipeline scripts, and no raw third-party data is redistributed.

## 5 Distillation and Student Fleet

### TEACHER SELECTION AND DISTILLATION LABELING

The training corpus arrived without labels in the target taxonomy, so a teacher model supplied them. Knowledge distillation (Hinton et al., 2015) provides the framework: a strong model produces the supervision signal that a smaller student is then trained to reproduce. The teacher was selected by a comparative run of candidate guards against the gold benchmark rather than by reputation. Llama Guard 3 8B (Inan et al., 2023) was chosen on that evidence, having obtained the highest macro-F1 (0.517) and the lowest benign false-positive rate (0.068) among the candidates, and having produced labels for the sparse self-harm and sensitive-information categories that the 1-billion-parameter variant largely failed to emit. The smaller variant was retained as a backup labeler for the override cascade described below.

Labeling ran one call per row, with the model loaded in 4-bit precision to fit an 8 GB development graphics processing unit, following the quantized-base approach of Dettmers et al. (2023), and with prompts truncated at 512 tokens. At a batch size of 12 the process sustained 3.4 rows per second, so labeling the full corpus of roughly 97,000 prompts took approximately 7.9 hours. The teacher emits hazard codes in its own taxonomy, which were mapped to the seven categories through the correspondence of Table 1, together with a binary safe or unsafe verdict. The resulting label became the authoritative training target, while the pre-teacher class hypothesis carried in the schema was kept only to audit agreement between the source's own labels and the teacher's.

One methodological constraint governs the whole design and deserves explicit statement. Distillation labeling touched the training buckets only. The gold benchmark was labeled independently, at the data stage, and no teacher ever ran over it. Had the benchmark been teacher-labeled, every reported score would measure agreement with the teacher rather than agreement with the ground truth, and the distillation gap this paper reports would be undefined.

### LABEL OVERRIDES

The teacher exhibits two reproducible blind spots that the corpus makes visible, and both were corrected by overrides applied in memory by the training loader, leaving the labels stored on disk as pure teacher output. Keeping the correction separate from the stored labels serves two purposes: the teacher's behavior remains auditable, and the overrides can be switched off, which is what makes them measurable in the ablations reported later. The first override addresses the synthetic corpus of personally identifiable information, which is such by construction while the teacher under-fires the corresponding category; the sensitive-information code was unioned onto the label of its 1,711 rows. The second addresses the two adversarial behavior suites, which are harmful by construction, so a safe verdict on them is necessarily wrong. Rows the teacher marked safe entered a cascade that consults the backup labeler, then the pre-teacher hypothesis, and finally drops the row if neither resolves it. That cascade recovered roughly 140 rows through the backup labeler and 11 through the hypothesis, and dropped 90. After overrides the deployable recipe holds 75,022 trainable rows and the research recipe 97,354.

The label distribution that results is unbalanced in two ways that shape every later decision. Safe rows account for 67.9% of the corpus, and among the unsafe categories self-harm is the scarcest at roughly 1,300 rows, which makes per-class rebalancing during training unavoidable rather than optional. The distribution is also effectively single-label: only 13 of the 25,997 unsafe labels in the commercial bucket carry two codes, that is 0.05%, and the noncommercial bucket carries none. Category co-occurrence therefore cannot be learned from this supervision, a limitation examined in Section 9.

## STUDENT FLEET AND TRAINING

The students span four architectural families, from a lexical baseline to a small generative model, and all of them predict the same target: a seven-dimensional multi-label vector in a fixed category order, where the all-zeros vector denotes a safe verdict. No eighth output is dedicated to the safe class, so safety is the absence of a violation rather than a competing category. The generative student emits its verdict as text and a shared contract decodes it back into the same vector, which keeps every model comparable under one scoring harness. Table 3 summarizes the fleet.

**Table 3.** The student fleet. All models predict the same seven-dimensional multi-label target.

| Model | Family | Parameters | Training configuration |
|---|---|---:|---|
| Floor (TF-IDF + LR) | lexical baseline | not applicable | word 1-2 and character 3-5 n-grams, 200,000 features, one-vs-rest logistic regression, balanced class weights |
| TextCNN | shallow network | ~4M | randomly initialized embeddings, seven sigmoid outputs |
| BiLSTM | shallow network | ~4M | randomly initialized embeddings, seven sigmoid outputs |
| MiniLM-L6 | encoder | 22M | seven sigmoid outputs over the pooled encoder |
| DistilBERT | encoder | 66M | seven sigmoid outputs over the pooled encoder |
| Qwen2.5-0.5B + LoRA | generative | 0.5B | low-rank adapter of 8.8M parameters (1.75%), maximum length 128, batch 8 with 2 accumulation steps, one epoch |

*Source: the authors.*

The lexical floor exists to set a threshold of usefulness: any learned model that fails to beat term-frequency features with logistic regression has not earned its complexity. The two shallow networks follow the convolutional text classifier of Kim (2014) and its recurrent counterpart, trained from randomly initialized embeddings. The two encoders are compressed members of the BERT family (Devlin et al., 2019), namely DistilBERT (Sanh et al., 2019) and MiniLM (Wang et al., 2020), each reduced to a single forward pass followed by seven sigmoid outputs. The generative student adapts a 0.5-billion-parameter instruction model through low-rank adaptation (Hu et al., 2021), training an adapter of 8.8 million parameters, or 1.75% of the base, which can subsequently be merged into the base weights.

Training used binary cross-entropy with exactly one rebalancing mechanism, never stacked. For the neural students, each category received a positive-class weight set to the square root of the ratio between its negative and positive counts, capped at 12. The square root damps the correction and the cap prevents the scarcest category from dominating the objective, which would trade overall quality for a single class. The lexical floor uses the balanced class weighting of its own implementation, a different mechanism serving the same purpose. All reported results use a fixed decision threshold of 0.5 for every category and every model, so that any two scores can be compared directly.

Finally, the generative student was also evaluated in a quantized form, because a 0.5-billion-parameter model in full precision is not the artifact that would run on a modest CPU. The adapter was merged into the base weights in 32-bit precision, and the merged model was converted and quantized through llama.cpp (Gerganov et al., 2023) at a pinned build, following established post-training quantization practice (Frantar et al., 2022; Lin et al., 2023). The 8-bit quantization is the variant reported in the latency analysis of Section 8, and its scores are reported separately from the full-precision figures, since the two are different runtimes and differ slightly.

## 6 Evaluation Protocol

Every model reported in this paper, whether teacher, student, or lexical baseline, was judged under one protocol: the same gold rows, the same metrics, and the same decision threshold. Uniform treatment is what allows the comparison between a 8-billion-parameter guard and a 4-million-parameter network to mean anything, and it is also what forbids the common practice of tuning each model to its own best operating point before comparing.

The gold benchmark divides into four slices, defined by the source of each row rather than by any post-hoc assignment, so that the division is reproducible from the data alone. Table 4 lists them with the role each one plays.

**Table 4.** The four evaluation slices, with scorable row counts.

| Slice | Sources | What it measures | Rows |
|---|---|---|---|
| dev | AttaQ, SALAD-Bench, SimpleSafetyTests | development and model selection | 1,585 |
| headline | AILuminate v1.0 (public) | the reference number, balanced and standard-aligned | 1,176 |
| stress | ALERT | robustness to adversarial phrasing | 3,000 |
| benign | reserved harmless prompts | false alarms, that is over-defense | 600 |
| **Total** | | | **6,361** |

*Source: the authors, from the gold benchmark composition.*

The headline slice carries the reference number because it is the only balanced slice, holding on the order of 100 to 300 rows per category, and because it is aligned to the public standard, which makes its value comparable in principle to figures reported elsewhere on the same benchmark. The stress slice is adversarial by construction and answers a different question, namely whether detection survives hostile phrasing. The benign slice is the only slice containing harmless prompts and therefore the only one that can measure false alarms; the other three are unsafe by construction and would report a false-positive rate of zero by definition.

The decision threshold was fixed at 0.5 for every category and every model. Per-category threshold tuning is implemented but was deliberately left off for all reported numbers, for a reason specific to this benchmark: the development slice contains no harmless rows, so tuning against it carries no penalty for false alarms and would systematically push thresholds down, improving detection while inflating over-defense. A fixed threshold avoids buying recall with an unmeasured cost.

Rows that could not be scored were counted rather than quietly discarded. Of the 6,802 gold rows, 441 carry a label that maps to no category in the taxonomy: 24 rows of the headline source fall under the elections hazard that Section 3 places out of scope, and 417 rows of one development source carry categories in a taxonomy that does not map onto the seven used here. The scorable set is therefore 6,361 rows, and that is the denominator of every reported metric. The gold labels are also effectively single-label, with no row carrying two categories, which means the diagonal of a class-by-class confusion matrix coincides with per-class recall and that a multi-label prediction cannot be rewarded by this benchmark.

Predictions and ground truth were compared as two binary matrices of dimensions n by 7. For each category, precision is the number of true positives divided by the sum of true and false positives, recall is the number of true positives divided by the sum of true positives and false negatives, and F1 is their harmonic mean, that is twice the product divided by the sum. The primary metric is macro-F1, the unweighted mean of the seven per-category F1 scores, which counts a rare category exactly as much as a frequent one and therefore reflects performance on the sparse categories that a volume-weighted average would conceal. Micro-F1, which pools the counts before computing the ratio, is dominated by the high-volume categories and is reported only where that pooled view is the question. Two further quantities carry the deployment reading: a binary confusion matrix in which any category firing above threshold counts as a block, and the false-positive rate on the 600 harmless rows, for which lower is better.

All reported point estimates carry a 95% percentile bootstrap confidence interval computed by resampling the evaluation rows 1,000 times under a fixed seed. The intervals are not decoration. Because macro-F1 weights the sparse categories equally, its sampling variance is dominated by them, and several apparent differences between the strongest models fall inside the resulting intervals. Claims in the following sections are stated at the resolution the intervals support, and no ordering is asserted where they overlap.

One element of the protocol remains unquantified and is reported as such. The gold labels were produced by mapping each benchmark's native categories onto the seven used here, and the fidelity of that mapping, meaning the agreement between the mapped label and independent human judgment, has not been measured. Every score in this paper is therefore agreement with the mapped gold, and a systematic error in the mapping would propagate to all models alike, teacher and students together, without changing their relative ordering.

## 7 Results

### THE DISTILLATION GAP

Table 5 reports every model on the three unsafe slices and on the benign slice, each figure with its 95% bootstrap confidence interval. Figure 2 shows the same macro-F1 values graphically, and Figure 3 isolates the benign false-positive rate.

**Table 5.** Teachers and student fleet on the gold benchmark. Macro-F1 per slice and benign false-positive rate, each with a 95% percentile bootstrap confidence interval (B = 1000), at a fixed threshold of 0.5.

| Model | Role | Params | dev | headline | stress | benign FPR |
|---|---|---:|---|---|---|---|
| Llama Guard 3 8B | teacher | 8B | 0.545 [.502,.583] | **0.556 [.524,.586]** | 0.588 [.566,.610] | 0.048 [.032,.065] |
| Llama Guard 3 1B | teacher | 1B | 0.522 [.478,.561] | 0.545 [.512,.575] | 0.598 [.575,.618] | 0.115 [.090,.140] |
| Qwen2.5-0.5B LoRA | student | 0.5B | 0.508 [.465,.542] | 0.503 [.472,.535] | 0.568 [.545,.590] | **0.038 [.023,.053]** |
| DistilBERT | student | 66M | 0.525 [.482,.562] | 0.500 [.468,.529] | **0.618 [.594,.641]** | 0.103 [.080,.128] |
| MiniLM-L6 | student | 22M | 0.532 [.491,.565] | 0.499 [.468,.527] | 0.596 [.572,.618] | 0.207 [.175,.238] |
| Floor (TF-IDF + LR) | student | not applicable | 0.494 [.450,.528] | 0.529 [.499,.553] | 0.533 [.509,.555] | 0.252 [.217,.285] |
| BiLSTM | student | ~4M | 0.451 [.406,.487] | 0.439 [.409,.468] | 0.477 [.452,.502] | 0.203 [.172,.240] |
| TextCNN | student | ~4M | 0.435 [.387,.471] | 0.451 [.420,.479] | 0.502 [.475,.526] | 0.208 [.177,.245] |

*Source: the authors.*

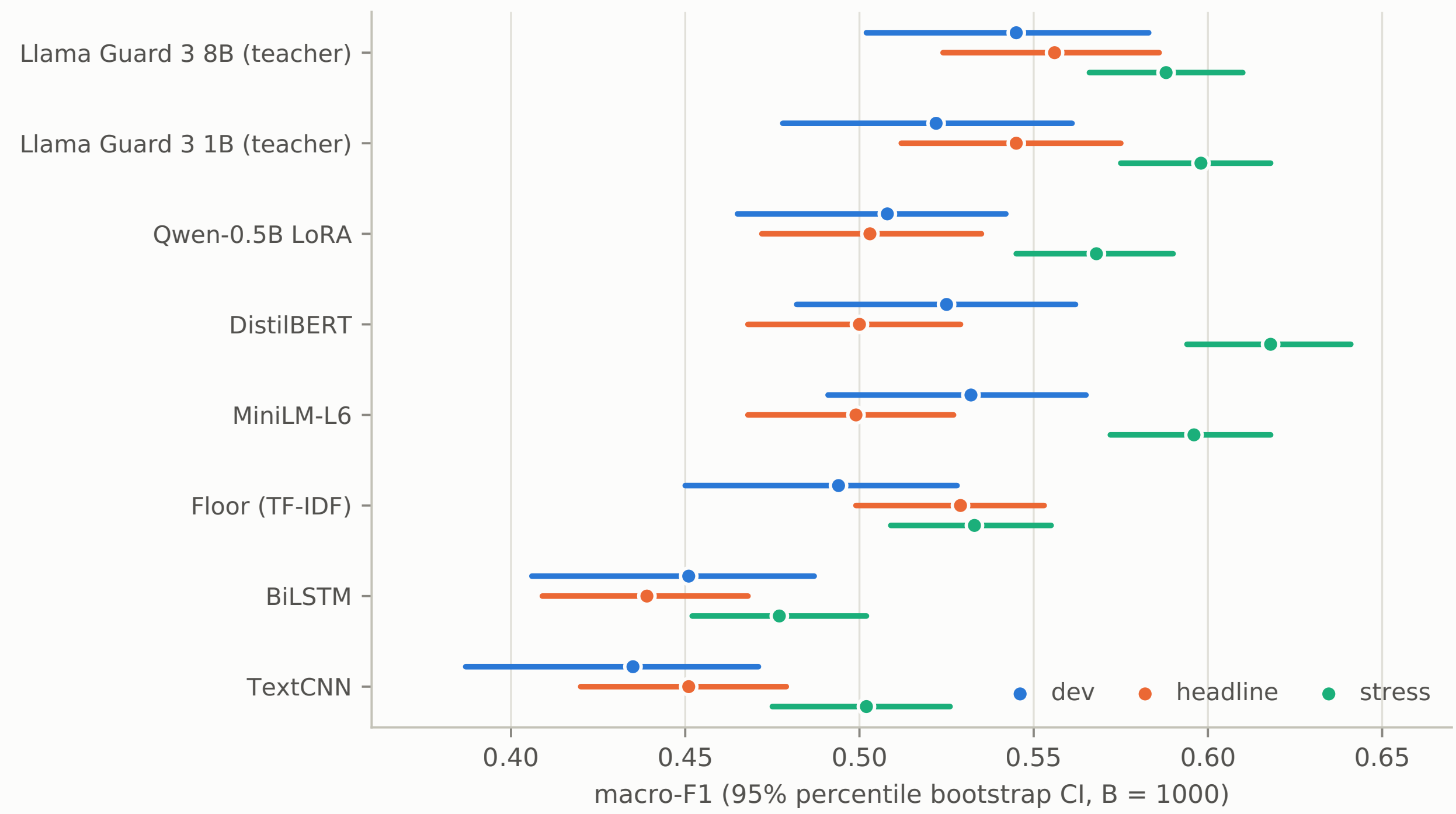

**Figure 2.** Macro-F1 by slice for every model, with 95% percentile bootstrap confidence intervals. *Source: the authors, from Table 5.*

Three readings follow, and the confidence intervals determine how strongly each may be stated. On the clean headline slice the teachers lead: the 8-billion-parameter teacher reaches 0.556 and the best student, the lexical floor, reaches 0.529, with partially overlapping intervals. On the adversarial stress slice the ordering inverts at the point estimate, since DistilBERT attains 0.618 against 0.598 for the smaller teacher, but the intervals overlap and the difference therefore falls inside sampling noise. The defensible statement is that the encoder matches the teachers on adversarial text, not that it beats them.

Over-defense is the one axis where the separation is real. Figure 3 shows the generative student at a benign false-positive rate of 0.038 with an interval of [.023,.053], entirely below the 0.115 of the smaller teacher, and below the 0.048 of the larger one at the point estimate. The two encoders and the shallow networks are worse than both teachers on this axis, and the lexical floor is the worst at 0.252, which is the expected failure of a purely lexical model confronted with prompts that merely look unsafe.

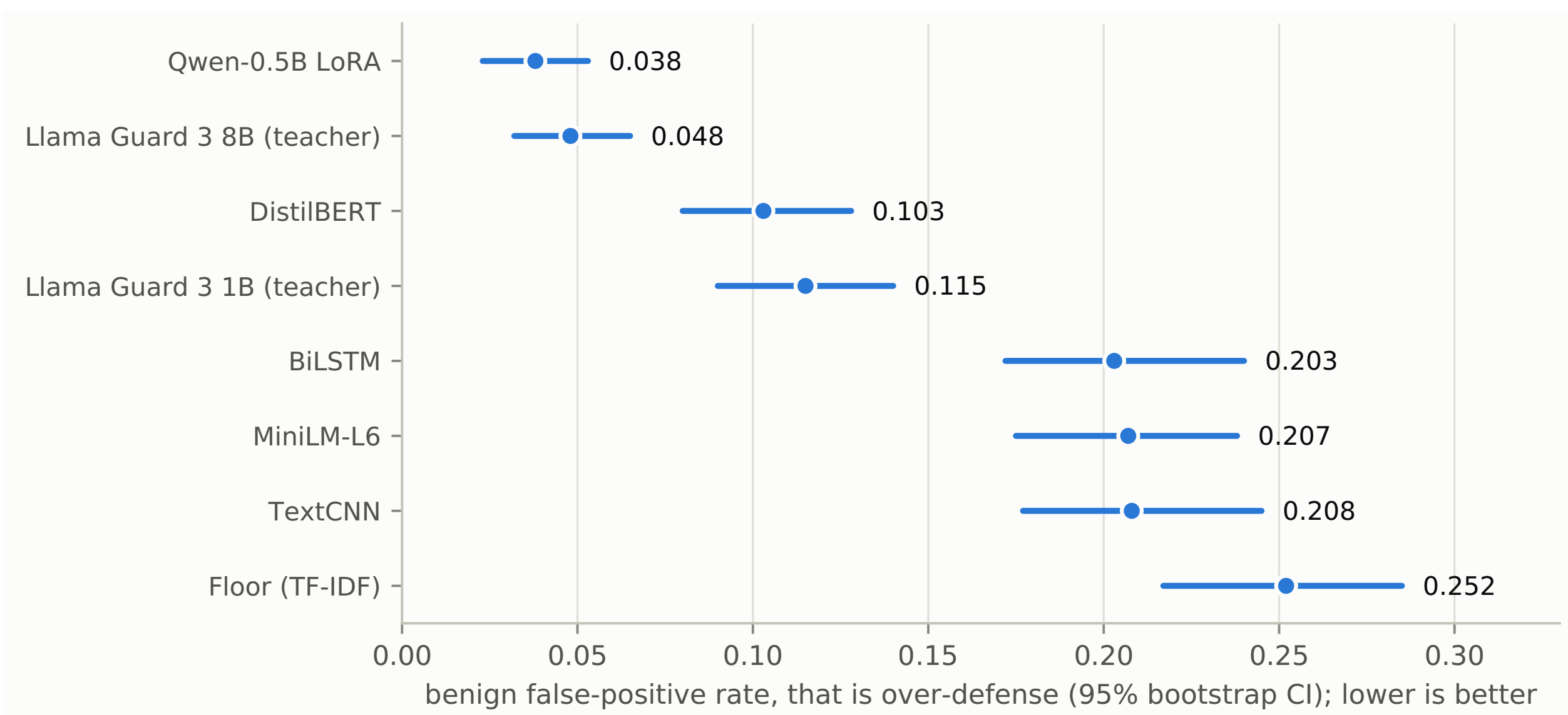

**Figure 3.** Benign false-positive rate, that is over-defense, with 95% bootstrap confidence intervals. Lower is better. *Source: the authors, from Table 5.*

Taken together, distillation transferred the teacher's signal into models between 16 and roughly 2,000 times smaller in parameter count, which match the teachers on adversarial text and, in the case of the generative student, reduce false alarms on harmless prompts. No global claim of superiority over Llama Guard 3 is supported by these intervals, and none is made.

### PER-CATEGORY BEHAVIOR

The macro-F1 values above are means over seven categories that behave very differently. Table 6 gives the per-category F1 on the balanced headline slice, and Figure 4 shows the same values.

**Table 6.** F1 per category on the headline slice. The macro column is the unweighted mean of the seven.

| Model | C1 | C2 | C3 | C4 | C5 | C6 | C7 | macro |
|---|---|---|---|---|---|---|---|---|
| Llama Guard 3 8B | 0.775 | 0.513 | 0.734 | 0.684 | 0.389 | 0.416 | 0.384 | 0.556 |
| Llama Guard 3 1B | 0.610 | 0.634 | 0.762 | 0.616 | 0.417 | 0.407 | 0.366 | 0.545 |
| Qwen2.5-0.5B LoRA | 0.698 | 0.463 | 0.631 | 0.542 | 0.416 | 0.419 | 0.354 | 0.503 |
| DistilBERT | 0.703 | 0.482 | 0.646 | 0.671 | 0.366 | 0.309 | 0.323 | 0.500 |
| MiniLM-L6 | 0.668 | 0.480 | 0.664 | 0.663 | 0.424 | 0.294 | 0.296 | 0.499 |
| Gold support | 200 | 200 | 300 | 100 | 200 | 76 | 100 | 1,176 |

*Source: the authors.*

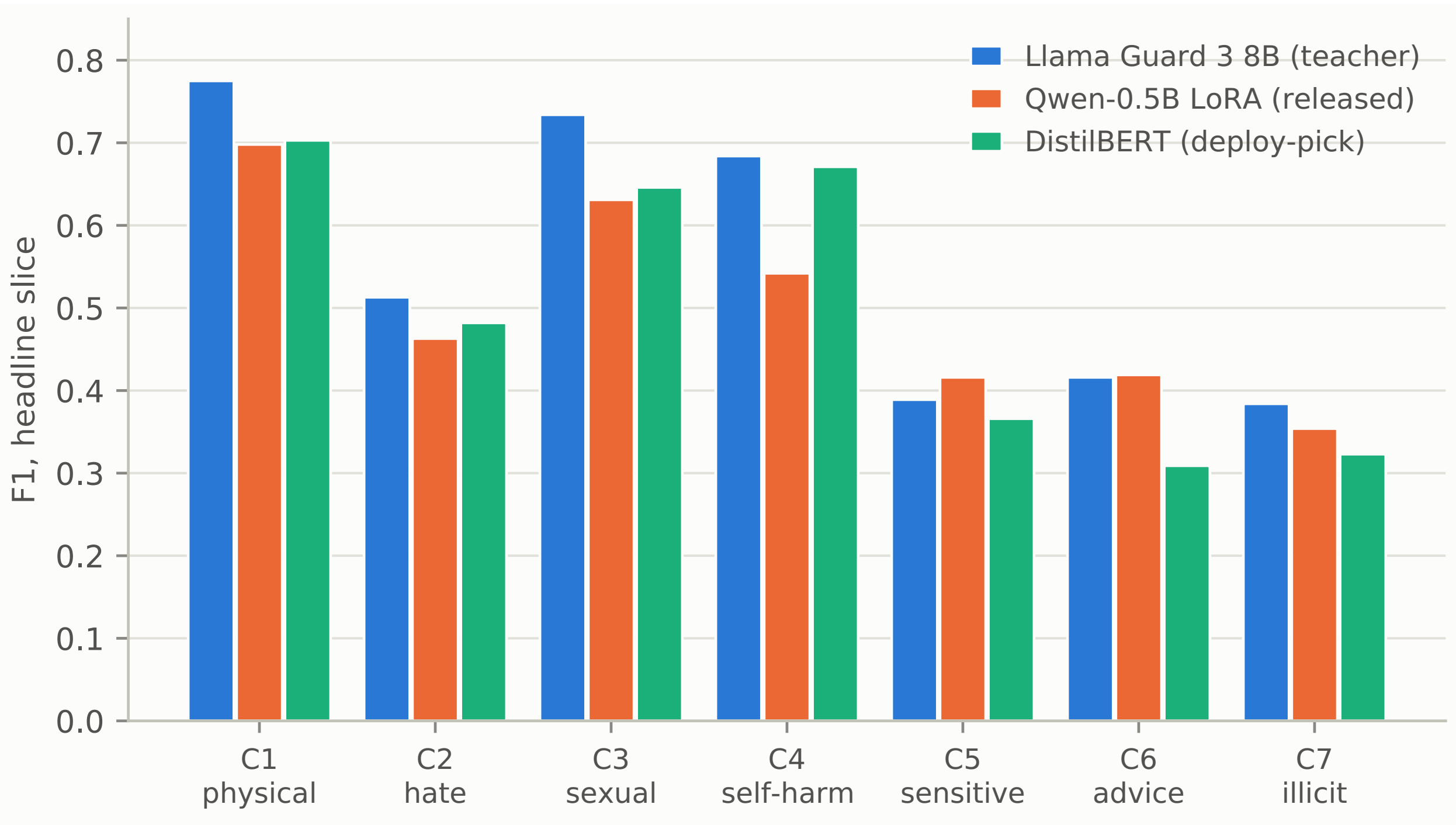

**Figure 4.** F1 per category on the headline slice, for the labeling teacher, the generative student and the encoder. *Source: the authors, from Table 6.*

Physical harm and sexual content are the categories every stronger model detects reliably, with F1 between 0.63 and 0.78. Sensitive information sits near 0.40 for all models. Harmful advice and illicit acts are the weakest on this slice, between 0.29 and 0.42, and the pattern holds for the teachers as much as for the students. Because macro-F1 counts each category equally, these three weak categories are what hold the reported means near 0.50 rather than any deficiency of the small models specifically. The uniformity of the failure across architectures and across two orders of magnitude in scale is itself the informative result, and Section 9 argues that it points at the category definitions rather than at model capacity.

### THE BLOCK OR PASS DECISION

Macro-F1 answers which category fired; a deployment also needs to know how often harmful text is stopped at all. Collapsing the seven outputs into a single binary decision, where any category above threshold counts as a block, yields Table 7.

**Table 7.** Binary block-or-pass confusion on the full scorable benchmark (600 harmless and 5,761 harmful rows).

| Model | TN | FP (over-defense) | FN (leaked) | TP | FPR | Unsafe recall | Unsafe precision |
|---|---|---|---|---|---|---|---|
| Qwen2.5-0.5B LoRA | 577 | 23 | 1,683 | 4,078 | 0.038 | 0.708 | 0.994 |
| DistilBERT | 538 | 62 | 1,319 | 4,442 | 0.103 | 0.771 | 0.986 |
| Llama Guard 3 8B | 571 | 29 | 1,520 | 4,241 | 0.048 | 0.736 | 0.993 |

*Source: the authors.*

On this binary view the generative student blocks 70.8% of harmful prompts against 73.6% for the teacher it was distilled from, while raising fewer false alarms, and the encoder blocks the most at 77.1% at the cost of a higher false-positive rate. Unsafe precision exceeds 0.98 for all three, so almost nothing that is blocked is harmless. The uncomfortable figure is the other side of the same table: every model, the 8-billion-parameter teacher included, lets roughly a quarter of the harmful prompts through. That is a property of an adversarial benchmark evaluated at a fixed threshold, not a defect peculiar to the distilled students, and it is the honest ceiling of the current recipe.

Figure 5 decomposes those errors by category. Since the gold labels are single-label, the diagonal is per-class recall, and the dominant off-diagonal cell for every model is the leftmost column: misses leak as safe rather than being assigned to a neighboring category. Hate and defamation is the category most often missed this way, followed by illicit acts.

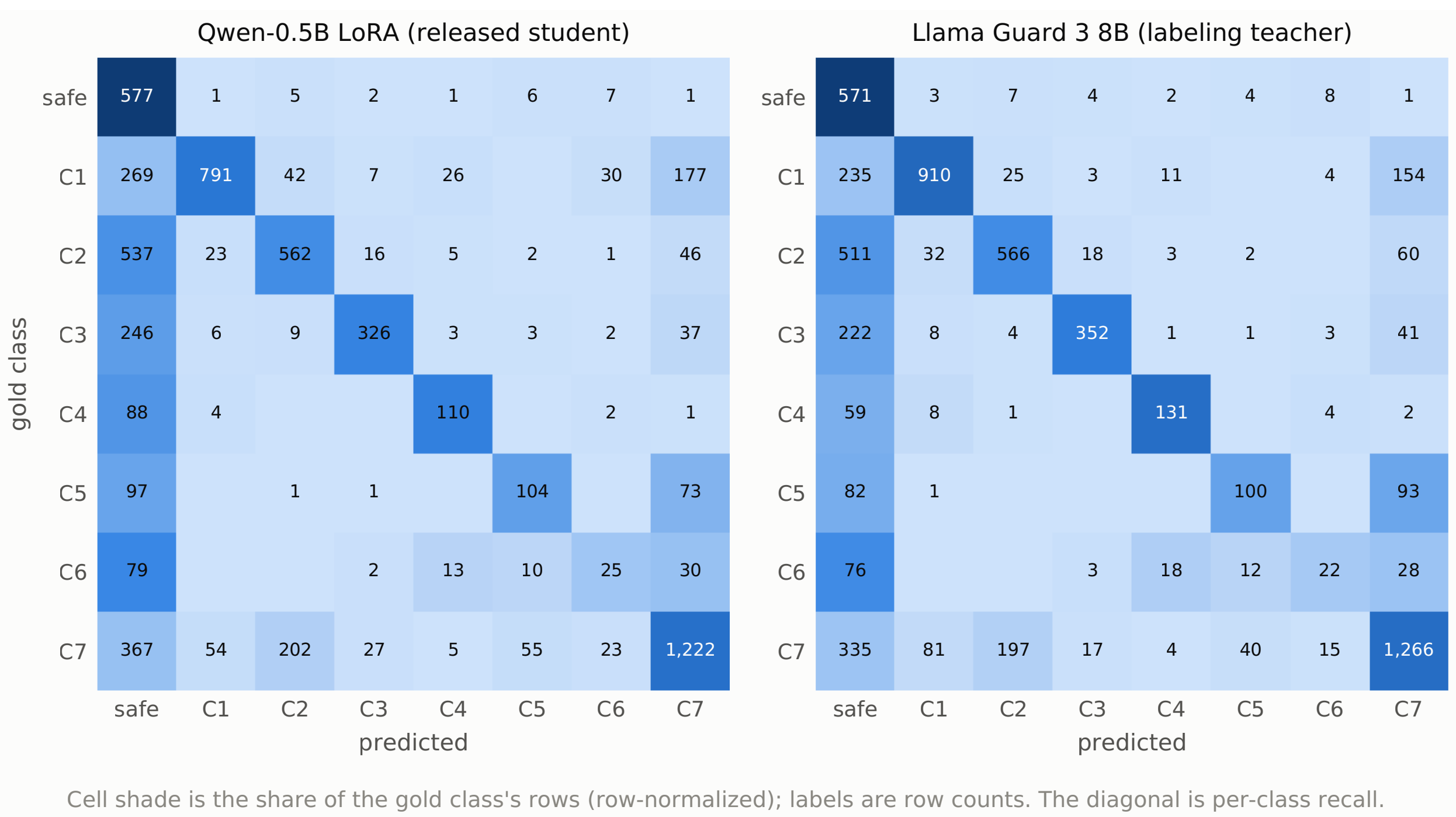

Cell shade is the share of the gold class's rows (row-normalized); labels are row counts. The diagonal is per-class recall.

**Figure 5.** Class confusion, gold against predicted, for the generative student and the labeling teacher. Shading is row-normalized; labels are row counts. *Source: the authors.*

### ABLATIONS

Which ingredient of the recipe actually carries the result was tested one factor at a time on the cheapest learned student, holding the seed fixed and changing a single lever per arm. Table 8 reports the outcome.

**Table 8.** One-factor-at-a-time ablations of the recipe, on the TextCNN student with a fixed seed. Deltas are against the recipe arm.

| Arm | Train rows | headline macro-F1 | stress macro-F1 | benign FPR | C4 F1 | Unsafe recall |
|---|---|---|---|---|---|---|
| recipe | 75,022 | 0.422 | 0.534 | 0.167 | 0.573 | 0.487 |
| without overrides | 75,112 | 0.443 (+0.021) | 0.521 (−0.013) | 0.173 (+0.007) | 0.623 (+0.049) | 0.478 (−0.009) |
| with noncommercial data | 97,354 | 0.427 (+0.005) | 0.528 (−0.006) | 0.193 (+0.027) | 0.645 (+0.071) | 0.500 (+0.013) |
| without rebalancer | 75,022 | 0.251 (−0.171) | 0.391 (−0.143) | 0.090 (−0.077) | 0.196 (−0.377) | 0.260 (−0.227) |

*Source: the authors.*

The rebalancer is the one decisive lever. Removing the per-class positive weighting costs 0.171 macro-F1 on the headline slice and 0.377 on the sparse self-harm category, and drives unsafe recall down by 0.227. The apparent improvement in false-positive rate, which falls by 0.077, is not a quality gain: without per-class weighting the model defaults to predicting safe, so it fires less often on everything, harmful text included. This delta is far outside the sampling noise the confidence intervals quantify, so the conclusion is robust.

The other two levers are within noise and must be reported as such. Removing the label overrides slightly improves headline macro-F1, by 0.021, while slightly worsening adversarial performance and false alarms. The overrides are therefore a correctness fix rather than a score improvement, and the useful finding is that correctness costs nothing measurable in F1. Adding the noncommercial data moves headline macro-F1 by only 0.005 while worsening the false-positive rate by 0.027, its one clear gain being 0.071 on the sparse self-harm category. Excluding the noncommercial bucket to keep the recipe deployable therefore costs almost nothing in headline quality and is in fact better on over-defense, which makes the license boundary a nearly free constraint rather than a sacrifice. All three of these deltas come from a single seed on the cheapest architecture, so their directions are reported and their magnitudes are not load-bearing.

## 8 Latency on CPU

Classification quality is only half of the question this work set out to answer, since the motivation was a safety layer that runs on commodity CPU hardware. Latency was therefore measured rather than assumed. All figures below come from single-request inference, that is a batch of one, on warm models, over 120 prompts sampled from the benchmark and stratified by length, with three timed passes. The host was a six-core, twelve-thread AMD Ryzen 5 5600X with 16 GB of memory. Two profiles were measured: a desktop profile using all available threads, and a constrained profile pinned to two virtual cores with the thread pools capped accordingly, which stands in for a small server allocation. Table 9 reports the median latency per request.

**Table 9.** Median (p50) CPU latency per request, in milliseconds, under two thread profiles.

| Model | Parameters | p50 desktop (ms) | p50 two vCPU (ms) |
|---|---|---|---|
| TextCNN | ~4M | 0.39 | 0.28 |

| Model | Parameters | p50 desktop (ms) | p50 two vCPU (ms) |
|---|---|---|---|
| Floor (TF-IDF + LR) | not applicable | 1.10 | 1.06 |
| BiLSTM | ~4M | 1.34 | 1.08 |
| MiniLM-L6 | 22M | 7.62 | 8.31 |
| DistilBERT | 66M | 24.49 | 27.96 |
| Qwen2.5-0.5B, 8-bit quantized | 0.5B | 127.52 | 178.56 |

*Source: the authors.*

The first result is that the whole small fleet is comfortably interactive on CPU. Every model except the generative student answers in under 28 ms at the median, even when restricted to two virtual cores, and the shallow and lexical models answer in around one millisecond or less. The generative student is an order of magnitude slower at roughly 128 ms on the desktop profile and 179 ms when constrained, which is still well below the second scale that a multi-billion-parameter guard requires on the same class of hardware. The teachers were not benchmarked under these profiles, since running an 8-billion-parameter model on two virtual cores is not a configuration this work proposes; the relevant comparison is that the entire distilled fleet operates in the millisecond range on hardware where such a guard would not be deployed at all.

The second result concerns the constrained profile. For the small models, capping the process at two virtual cores changes latency very little, and for TextCNN and BiLSTM it is marginally faster. Single-request inference on a light model is bound by memory access and framework overhead rather than by arithmetic throughput, so additional threads have little to contribute and their coordination cost can dominate. Thread count begins to matter only for the generative student, whose latency rises by roughly 40% under the cap, since it is the only model in the fleet doing enough computation per request to use the extra cores.

Figure 1 places latency and quality on the same axes and shows the resulting trade-off surface.

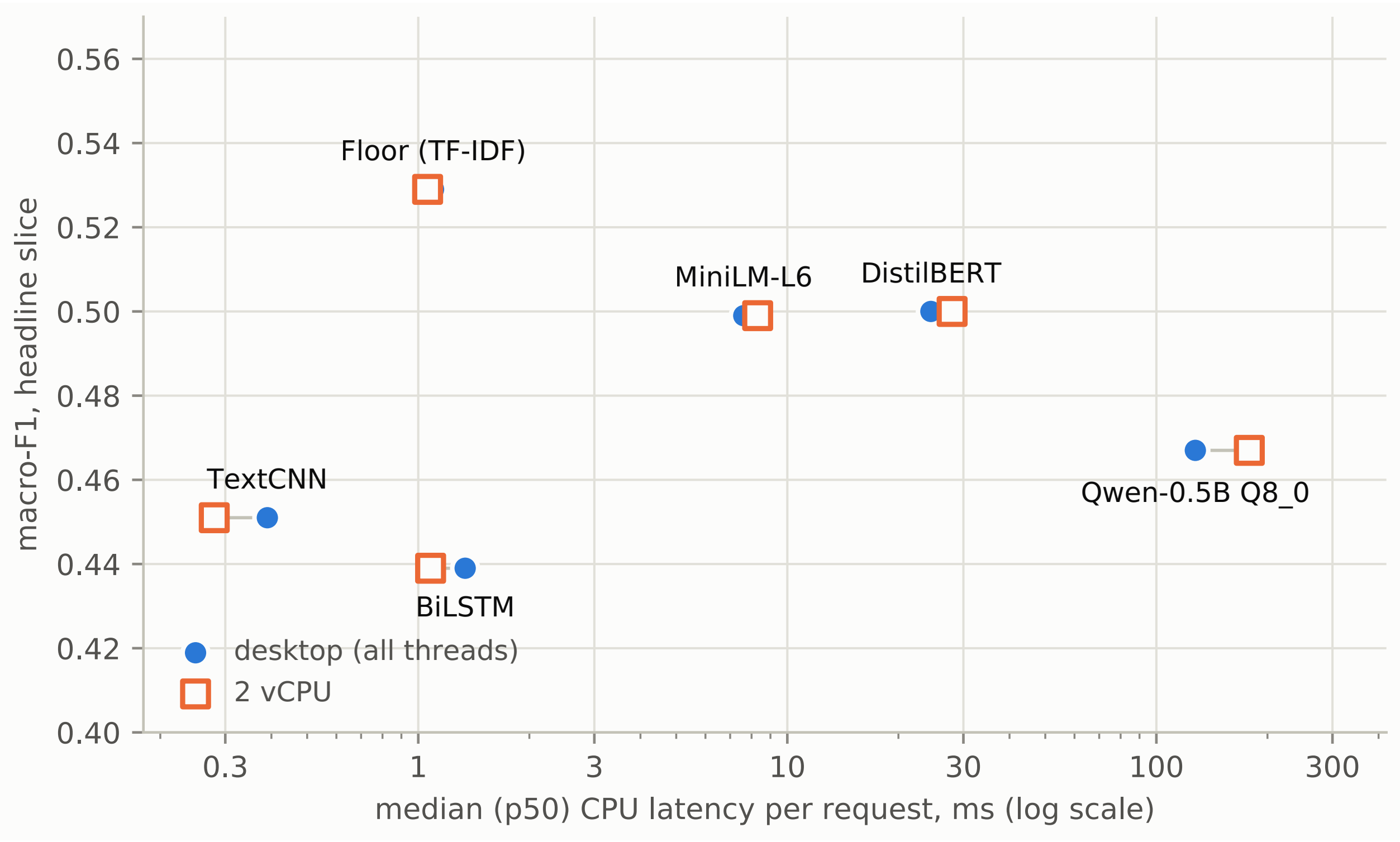

**Figure 1.** Median CPU latency, on a logarithmic scale, against macro-F1 on the headline slice, for both thread profiles. *Source: the authors, from Tables 5 and 9.*

The figure makes one comparison unavoidable. The encoder is not merely cheaper than the generative student, it is better on the axes that matter here: DistilBERT answers roughly five times faster and attains higher adversarial macro-F1 (0.618 against 0.568 in Table 5). On the evidence of this benchmark, the encoder is the optimum of the fleet, and a generative student earns its place only where the surrounding stack is already generative, so that a single serving path can carry both text generation and classification. That is an integration argument rather than a quality argument, and the two should not be conflated.

One accounting note is required to read Figure 1 against Table 5 without confusion. The generative student appears in Table 5 at a headline macro-F1 of 0.503, measured in full precision, whereas Figure 1 plots it at 0.467, the value obtained from the 8-bit quantized model whose latency is the one reported here. Quality figures are always paired with the runtime that produced them, and the difference of roughly 0.04 between the two is consistent with the borderline-decision shift that quantization introduces at a fixed threshold. Both values are legitimate; they describe different artifacts, and reporting the quantized figure alongside the quantized latency is what keeps the comparison honest.

## 9 Discussion and Limitations

### WHAT THE RECIPE BUYS

The central practical finding is that the license boundary is nearly free. Because the corpus is partitioned at that boundary and both recipes are otherwise identical, the cost of restricting training to commercially usable data is directly measurable, and it is small: adding the noncommercial bucket moves headline macro-F1 by 0.005 while worsening the false-positive rate on harmless prompts by 0.027. A recipe that many would treat as a compromise turns out to be the better one on over-defense, and the only quantity it clearly gives up is support for the sparse self-harm category. This is the kind of claim that is only available to a pipeline built to make it, and it argues for treating license partitioning as a design parameter to be measured rather than a constraint to be lamented.

The second finding concerns which question is being asked of the classifier. Macro-F1 near 0.50 and a binary block rate near 0.71 for the same model are not in tension: they measure different things. Assigning the correct category among seven is substantially harder than deciding whether anything harmful is present, and the results show the coarse decision holding up well precisely where fine attribution is weakest. Unsafe precision above 0.98 across the fleet says that what these models block is almost never harmless. A system that needs a gate benefits from the strength of the coarse decision; a system that needs to explain which policy was violated inherits the weakness of the fine one.

The over-defense result deserves emphasis because it inverts the usual expectation. The smallest generative student produces the fewest false alarms of any model tested, teachers included, at a benign false-positive rate of 0.038 against 0.048 and 0.115 for the two teachers, and the separation from the smaller teacher is outside the confidence intervals. A plausible reading is that distillation with per-class rebalancing on a corpus two-thirds composed of harmless prompts teaches a sharper decision boundary near the benign region than in-context policy prompting produces, though this work did not run the ablation that would establish the mechanism.

### THE TAXONOMY IS THE LIMITING FACTOR, NOT MODEL SCALE

The per-category results in Table 6 point at a conclusion the paper did not set out to test. Harmful advice collapses for every model evaluated, and it collapses for the 8-billion-parameter teacher as thoroughly as for a 4-million-parameter network, at an F1 of roughly 0.20 with recall near 0.14 on the full benchmark. When a failure is invariant across two orders of magnitude of capacity and across four architectural families, the category rather than the model is the most economical explanation. The confusion structure supports that reading directly: when the gold label is harmful advice, models predominantly emit safe, because the advice does not read as harmful in isolation, or they emit illicit acts, because they detect the underlying act instead of its advisory framing. The category as defined here extends beyond the specialized-advice hazard of the public standard to cover harm vectored through advice, and it is exactly that extension which fails to separate linguistically from benign advice on one side and from the act itself on the other.

The corresponding revision is to define harmful advice strictly as the standard's specialized-advice hazard and to reassign advice to commit a given act to the category of that act. Advice-vectored harm behaves like a modifier of an act rather than a sibling category, and modeling it as its own label is what produces the collapse. A second and milder case is the boundary between violent and nonviolent crime, where the errors run in both directions and dominate the residual confusion of the strongest models. Here the evidence does not support creating a new category; it supports sharpening the annotation guidance at the boundary, since the distinction is real and the models simply have not been told where the line falls.

One further pattern has practical weight. Across every model, the largest off-diagonal mass in Figure 5 is the column of safe predictions: when a model errs on harmful text, it overwhelmingly fails to fire at all rather than firing on the wrong category. Misses therefore surface as silent passes rather than as mislabeled blocks, which means that raising sensitivity is the lever that addresses most of the residual error, at the cost in over-defense that the benign slice exists to quantify.

## LIMITATIONS

The teacher is a ceiling. Every student is trained toward one model's judgment, so its blind spots propagate, and the two overrides documented in Section 5 correct only the two that the corpus made visible. The teacher shootout was also bounded by available hardware: candidate guards larger than 8 billion parameters could not be run efficiently on the development graphics processing unit and were therefore excluded, so whether a larger teacher would yield a better student is untested rather than answered.

No head-to-head comparison against other open guards was possible. ShieldGemma (Zeng et al., 2024) and Granite Guardian (Padhi et al., 2024) organize their outputs around hazard taxonomies that do not map cleanly onto the seven categories used here, and every metric in this paper is defined against the standard-aligned scheme of Table 1. Forcing an approximate correspondence would produce numbers whose disagreements reflect the mapping rather than the models, so the comparison was left undone. This bounds the paper's positioning: the students are compared against the teachers they distill and against a lexical floor, not against the open guard landscape as a whole.

The multi-label formulation is, in this version, an interface rather than a demonstrated capability, and the measurements make the situation precise. Only 13 of the 25,997 unsafe labels in the commercial bucket carry two categories, or 0.05%, the noncommercial bucket carries none, and the gold benchmark carries none at all. The generative student consequently emits no multi-label verdict on the entire benchmark, reproducing the teacher's single-label habit exactly, while the sigmoid-headed students emit two or three categories on between 4.5% and 8.2% of their unsafe verdicts, since independent per-category thresholds can fire together and are not bound to the teacher's output format. Category co-occurrence therefore cannot be learned from this supervision and could not be scored by this benchmark even if it were: the scarcity is a property of the distillation signal, not evidence of truncated generation. Establishing genuine multi-label behavior requires supervision that assigns each category independently, which is a relabeling task rather than a training adjustment.

Statistical resolution is limited in two ways. The sparse categories carry thin gold support, between 76 and 200 rows on the balanced headline slice, which widens the macro-F1 confidence intervals enough that several differences among the strongest models overlap; sensitive information and self-harm are data-starved on the training side as well. The ablations, in turn, come from a single seed on the cheapest architecture, so only the rebalancer effect is beyond sampling noise, and the override and data-bucket effects are reported as directions rather than magnitudes.

Two further boundaries are worth stating plainly. The fidelity of the gold mapping was not measured, so all reported scores are agreement with a mapped label rather than with independent human judgment; an error in the mapping would affect all models alike without reordering them. And the work is monolingual: every training and evaluation row is English, the Brazilian-Portuguese resources available are under noncommercial terms and were reserved for future work, so no multilingual claim is made or supported here. Finally, at the fixed threshold this protocol uses, roughly a quarter of harmful prompts pass every model tested, the teacher included, which is the honest performance ceiling of the current recipe rather than a property of distillation.

## 10 Conclusion

This paper set out to determine whether a safety classifier of useful quality can be produced under an auditable recipe and run on commodity CPU hardware, and it answers in the affirmative with the qualifications the evidence requires. A strong open guard labeled a corpus drawn from 24 public datasets into seven categories aligned to a public hazard standard, a fleet of small students distilled that signal, and all of them were scored against an independent gold benchmark of 6,361 rows that no teacher ever saw, across four slices including one composed entirely of harmless prompts.

Three results survive their confidence intervals. Distillation transfers the teacher's signal into models between 16 and roughly 2,000 times smaller in parameter count, which match the teachers on adversarial text rather than beating them, since the intervals overlap. Over-defense is where the small models genuinely win: the generative student raises false alarms on 3.8% of harmless prompts against 4.8% and 11.5% for the two teachers, a separation the intervals sustain. And the entire fleet except the generative student classifies in under 28 ms per request at the median on CPU, even when restricted to two virtual cores, which places the encoder rather than the generative model at the optimum of the quality and latency trade-off. Of the recipe's ingredients, only per-class rebalancing proved decisive; the license restriction that keeps the corpus commercially usable costs almost nothing and improves over-defense.

Two findings were not among the objectives and matter more than some that were. Restricting training data at the license boundary is measurable and nearly free, which turns an ordinarily unexamined constraint into a reportable quantity. And the ceiling on per-category quality is set by the taxonomy rather than by model capacity: the harmful-advice category collapses identically for a 4-million-parameter network and for an 8-billion-parameter teacher, which is evidence about the category definition and not about scale.

The work is presented as a resource and benchmark contribution. It makes no claim of state-of-the-art performance, and specifically no claim of superiority over the guards it distills from, since on the clean reference slice those guards remain ahead. Its value is that every step is reproducible from public sources, that the evaluation is independent of the supervision, and that the failure modes are reported at the resolution the data supports.

Four directions follow directly. The taxonomy should be revised to define harmful advice strictly as the standard's specialized-advice hazard, reassigning advice to commit an act to the category of that act, and to sharpen the guidance at the boundary between violent and nonviolent crime without splitting either category. The sparse categories need targeted data work, through paraphrase and generation rather than back-translation, which tends to sanitize the intent that makes a prompt harmful. Genuine multi-label behavior requires relabeling that scores each category independently, so that co-occurrence becomes both trainable and measurable. And the monolingual boundary should be crossed with resources whose licenses permit it, so that a Portuguese-language evaluation can be reported on the same footing as the English one presented here.

## References

The list below is generated from `literatura/referencias.bib`, the single source of truth for this article's citations, and holds 53 works, each verified at source before entry. Formatting follows APA 7; conversion to a venue's style is mechanical from the same file.

# A Appendix A: Reproducibility and Licensing

Every dataset used in this work is public and every derived table is regenerable from it. The corpus is rebuilt by running the per-source normalization scripts, which download each dataset from its own distribution channel and convert it to the common schema of Section 4; no raw third-party data is redistributed by this work. The evaluation numbers are likewise reproducible without rerunning any model, since per-row predictions are materialized once and every table in Sections 7 and 8 is computed from those files. The per-category table in particular is produced by a script that recomputes the per-slice macro-F1 from the same predictions and refuses to emit a table if those values fail to reproduce the reported ones, so a divergence between the two surfaces as an error rather than as a silent inconsistency. The figures are generated from the same numbers by a single script held with the article.

Licensing is handled structurally rather than by disclaimer. Every row carries the bucket its source's license places it in, so a model's license posture is fully described by the buckets it consumed. The deployable recipe consumes only the commercial bucket. Any model that consumed the noncommercial bucket is a research artifact and is reported as such, which is why the two recipes are always reported separately rather than merged into a single best number. Table A1 lists the full roster with the license that determined each assignment.

**Table A1.** Source roster with licenses and bucket assignment. The bucket follows the license; the evaluation bucket is never trained on.

| Dataset | License | Bucket |
|---|---|---|
| AdvBench | MIT | commercial |
| Anthropic HH-RLHF (harmless-base) | MIT | commercial |
| Anthropic Red Team | MIT | commercial |
| Aya Red-teaming | Apache 2.0 | commercial |
| CoCoNot | ODC-BY 1.0 (attribution) | commercial |
| DiaSafety | Apache 2.0 | commercial |
| DoNotAnswer | Apache 2.0 | commercial |
| Gretel synthetic PII (finance, multilingual) | Apache 2.0 | commercial |
| HarmBench | MIT | commercial |
| Jigsaw Toxic Comment | CC0 | commercial |
| MedSafetyBench | MIT | commercial |
| ProsocialDialog | CC BY 4.0 | commercial |
| UltraSafety | MIT | commercial |
| WildGuardMix | ODC-BY 1.0 (attribution) | commercial |
| WildJailbreak (vanilla portion) | ODC-BY 1.0 (attribution) | commercial |
| BeaverTails | CC BY-NC 4.0 | noncommercial |
| HateBR | research only | noncommercial |
| ToLD-BR | CC BY-SA 4.0 (data), MIT (code) | noncommercial |
| ToxicChat | CC BY-NC 4.0 | noncommercial |
| AILuminate v1.0 (public) | Apache 2.0 | evaluation |
| ALERT | CC BY-NC-SA 4.0 | evaluation |
| AttaQ | MIT | evaluation |
| SALAD-Bench | Apache 2.0 | evaluation |
| SimpleSafetyTests | CC BY 4.0 | evaluation |
| SafetyBench | MIT | normalized, excluded from v0 |

*Source: the authors, from the per-source license audit.*

Three entries in the table deserve a note. The adversarial stress benchmark carries noncommercial terms, which is unproblematic because it is used only for evaluation and never for training, and this is precisely the separation the bucket structure enforces. The Brazilian-Portuguese resources are the two the paper would need in order to make any multilingual claim, and both are restricted, which is why Section 9 reports the monolingual boundary as a licensing consequence rather than an oversight. SafetyBench was normalized into the schema and then excluded, because its multiple-choice format with a withheld answer key does not match the classification task; the folder is retained so that the decision is visible rather than implicit.

The labeled corpus is released as a dataset with the same four splits used here, so that the training and evaluation partitions can be inspected without rerunning the pipeline, with the noncommercial split kept separate to preserve the license boundary.

Two things are deliberately not part of this release. The gold benchmark's mapping fidelity, meaning the agreement between each benchmark's native categories and the seven used here, was not measured, so the reproducibility claim covers the mapping as applied and not its correctness against independent human judgment. And no model trained on the noncommercial bucket is offered for use beyond research, since the license does not permit it; the numbers from that recipe appear in this paper only to quantify what the license boundary costs.